# Non-rigid image registration using fully convolutional networks with deep self-supervision

Hongming Li, Yong Fan

Section for Biomedical Image Analysis (SBIA), Center for Biomedical Image Computing and Analytics (CBICA), Department of Radiology, Perelman School of Medicine, University of Pennsylvania, Philadelphia, PA, 19104, USA

**Abstract.** We propose a novel non-rigid image registration algorithm that is built upon fully convolutional networks (FCNs) to optimize and learn spatial transformations between pairs of images to be registered. Different from most existing deep learning based image registration methods that learn spatial transformations from training data with known corresponding spatial transformations, our method directly estimates spatial transformations between pairs of images by maximizing an image-wise similarity metric between fixed and deformed moving images, similar to conventional image registration algorithms. At the same time, our method also learns FCNs for encoding the spatial transformations at the same spatial resolution of images to be registered, rather than learning coarse-grained spatial transformation information. The image registration is implemented in a multi-resolution image registration framework to jointly optimize and learn spatial transformations and FCNs at different resolutions with deep self-supervision through typical feedforward and backpropagation computation. Since our method simultaneously optimizes and learns spatial transformations for the image registration, our method can be directly used to register a pair of images, and the registration of a set of images is also a training procedure for FCNs so that the trained FCNs can be directly adopted to register new images by feedforward computation of the learned FCNs without any optimization. The proposed method has been evaluated for registering 3D structural brain magnetic resonance (MR) images and obtained better performance than state-of-the-art image registration algorithms.

## Introduction

Medical image registration plays an important role in many medical image processing tasks [1, 2]. The image registration is typically formulated as an optimization problem to seek a spatial transformation that establishes pixel/voxel correspondence between a pair of fixed and moving images by maximizing a surrogate measure of the spatial correspondence between images, such as image intensity correlation between registered images. Since the image registration optimization problem is typically solved using iterative optimization algorithms, conventional image registration algorithms are often computational expensive and make the image registration time consuming.

Following the success of deep learning in a wide variety of computer vision and image processing tasks, such as image recognition, classification, and segmentation [3-6], several studies have leveraged deep learning techniques to improve medical image registration. In particular, deep learning techniques have been used to build prediction models of spatial transformations for achieving image registration under a supervised learning framework [7-10], besides learning image features for the image registration using stacked autoencoders [11]. Different from the conventional image registration algorithms, the deep learning based image registration algorithms formulate the image registration as a multi-output regression problem [7-10]. They are designed to predict spatial relationship between image pixel/voxels from a pair of images based on their image patches. The learned prediction model can then be applied to images pixel/voxel-wisely to achieve an overall image registration.

The prediction based image registration algorithms typically adopt convolutional neural networks (CNNs) to learn informative image features and a mapping between the learned image features and spatial transformations that register images in a training dataset [7-10]. Similar to most deep learning tasks, the quality of training data plays an important role in the prediction based image registration, and a variety of strategies have been proposed to build training data, specifically the spatial transformations that register images in a training dataset [7-10]. Particularly, synthetic deformation fields can be simulated and applied to a set of images to generate new images with known spatial transformations [7]. However, the synthetic deformation fields may not effectively capture spatial correspondences between real images. Spatial transformations that register pairs of images can also be estimated using conventional image registration algorithms [8, 9]. However, a prediction based image registration model built upon such a training dataset is limited to estimating spatial transformations captured by the adopted conventional image registration algorithms. The estimation of spatial transformations that register pairs of images can also be guided by shape matching [10]. However, a large dataset of medical images with manual segmentation labels is often not available.

Inspired by spatial transformer network (STN) [12], deep CNNs in conjunction with STNs have been proposed recently to learn prediction models for image registration from pairs of fixed and moving images in an unsupervised fashion [7, 13]. In particular, DirNet learns CNNs from pairs of images to estimate 16×16 2D control points of cubic B-splines for representing spatial transformations that register 2D images by optimizing an image similarity metric between fixed and transformed moving images [7]. Instead of representing spatial transformations using B-splines, ssEMnet estimates coarse-grained deformation fields at a low spatial resolution and uses bilinear interpolation to obtain dense spatial transformations for registering 2D images by optimizing an image similarity metric derived from a convolutional autoencoder between fixed and transformed moving images [13]. Although these methods have achieved promising image registration performance, their predicted coarse-grained spatial transformation measures often fail to characterize small deformations between images.

Building upon fully convolutional networks (FCNs) that facilitate voxel-to-voxel learning [3], we propose a novel deep learning based non-rigid image registration framework to optimize and learn spatial transformations between pairs of images to be registered. In particular, our method trains FCNs to estimate voxel-to-voxel spatial transformations for registering images by maximizing their image-wise similarity metric, similar to conventional image registration algorithms. To account for potential large deformations between images, a multi-resolution strategy is adopted to jointly optimize and learn spatial transformations at different resolutions. FCNs at higher resolutions are connected to lower-resolution FCNs using convolutional and pooling operators, while lower-resolution FCNs are connected to higher-resolution FCNs using deconvolutional operators. The image similarity measures between the fixed and deformed moving images are evaluated at different image resolutions to serve as deep self-supervision so that FCNs at different resolutions are jointly learned with a typical backpropagation based deep learning setting. Since our method simultaneously optimizes and learns spatial transformations for the image registration in an unsupervised fashion, the registration of a set of images is also a training procedure, and the trained networks can be directly adopted to register new images using feedforward computation. The proposed method has been evaluated based on 3D structural brain magnetic resonance (MR) images and obtained better performance than state-of-the-art image registration algorithms.

## Methods

### Image registration by optimizing an image similarity metric

Given a pair of fixed image $I_f$ and moving image $I_m$, the task of image registration is to seek a spatial transformation that establishes pixel/voxel-wise spatial correspondence between the two images. Since the spatial correspondence can be gauged with a surrogate measure, such as an image intensity similarity measure between the fixed and the transformed moving images, the image registration task can be formulated as an optimization problem to optimize a spatial transformation that maximizes the image similarity measure between the fixed image and the transformed moving image. For non-rigid image registration, the spatial transformation is often characterized by a dense deformation field $D_v$ that encodes displacement vectors between spatial coordinates of $I_f$ and their counterparts in $I_m$. For mono-modality image registration, mean squared intensity difference and normalized correlation coefficient are widely adopted image similarity measures. As the image registration problem is an ill-posed problem, regularization techniques are usually used to obtained smooth and physically plausible spatial transformations [1, 2]. In general, the optimization based image registration problem is formulated as

$$\min_{D_v} -S\left(I_f(v), I_m(D_v \circ v)\right) + \lambda R(D_v) \tag{1}$$

where $v$ is spatial coordinates of pixel/voxels in $I_f$, $D_v \circ v$ is deformed spatial coordinates of pixel/voxels by $D_v$ in $I_m$, $S(I_1, I_2)$ is an image similarity measure, $R(D)$ is a regularizer on the deformation field, and $\lambda$ controls the trade-off between the image similarity measure and the regularization on the spatial transformation.

The spatial transformation is typically regularized by square $L_2$-norm of its derivatives in order to obtain spatially smooth image registration results. However, it might over-penalize large deformations. To alleviate this problem, a total variation based regularizer can be used [14]

$$R(D_v) = \sum_{n=1}^{N} \|\nabla D_v(n)\|_1, \tag{2}$$

where $N$ is the number of pixel/voxels in the deformation field.

The above optimization problem can be solved by gradient descent based methods [1, 2]. However, such an optimization based image registration task is typically computational expensive and time consuming.

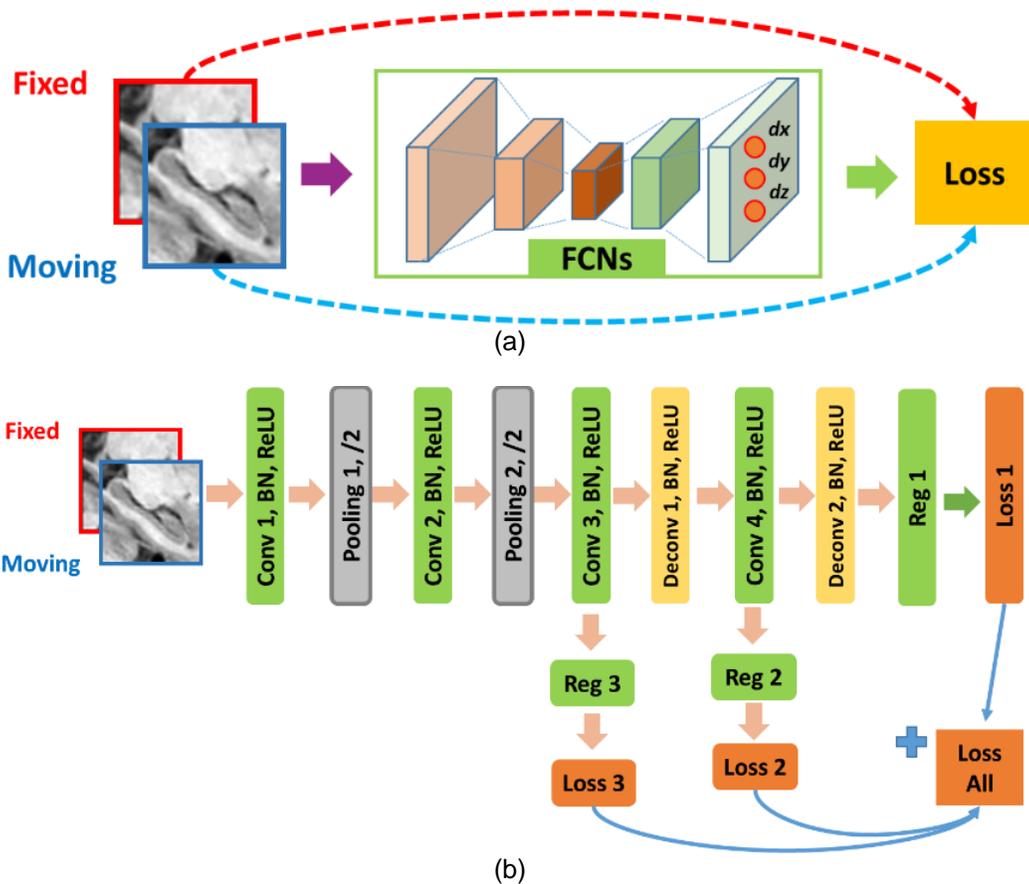

Fig. 1. Schematic illustration for image registration based on FCNs. (a) Overall architecture of the image registration framework, (b) an example FCN module for voxel-to-voxel multi-output regression of deformation fields in a multi-resolution image registration framework.

**Image registration using FCNs for optimizing an image similarity metric**

To solve the optimization problem for image registration, we build a deep learning model using FCNs to learn informative image feature representations and a mapping between feature representation and the spatial transformation between images at the same time. Particularly, we adopted FCNs to facilitate voxel-to-voxel learning so that we could directly obtain deformation fields at the same spatial resolution of the input images [3]. The registration framework of our method is illustrated by Fig.1a. In particular, each pair of fixed and moving images are concatenated as a two-channel input to the deep learning model for learning spatial transformations that optimize image similarity measures between the fixed and transformed moving images. The deep learning model consists of FCNs with de/convolutional (Conv) layers, batch normalization (BN) layers, activation (ReLU) layers, pooling layers, and multi-output regression layers, as illustrated by Fig. 1b. Each of the regression layer (Reg) is implemented as a convolutional layer whose output has the same size of the input images in spatial domain and multiple channels for encoding displacements in different spatial dimensions of the input images.

The pooling operation is usually adopted in CNNs to obtain translation-invariant features and increase reception fields of the CNNs, as well as to reduce the spatial size of the CNNs to reduce the number of parameters and computational cost [6]. However, the multi-output regression layers after pooling operations produce coarse outputs which have to be interpolated to generate deformation fields at the same spatial resolution of the input images [7, 13]. An alternative way to obtain fine-grained deformation fields is to stack multiple convolutional layers without any pooling layer. However, such a network architecture would have more parameters to be learned and decrease the efficiency of the whole network. In this study, we adopt deconvolutional operators for upsampling in a network with pooling layers [3], instead of choosing a specific interpolation scheme, such as cubic spline and bilinear interpolation [7, 13]. Such a network architecture also naturally leads to a multi-resolution image registration that has been widely adopted in conventional image registration algorithms [2].

In this study, normalized cross-correlation is used as the image similarity metric between images, and a total variation based regularizer as formulated by Eq. (2) is adopted to regularize the deformation fields. Therefore, the loss layer evaluates the registration loss between the fixed and deformed moving images as formulated by Eq. (1).

**Multi-resolution image registration with deep self-supervision**

Our multi-resolution image registration method is built upon FCNs with deep self-supervision, as illustrated by Fig. 1b. Particularly, the first 2 pooling layers in conjunction with their preceding convolutional layers progressively reduce the spatial size of the convolutional networks so that informative image features can be learned by the 3$^{rd}$ convolutional layer to predict voxel-wise displacement at the same spatial resolution of downsampled input images. And the subsequent deconvolutional layers learn informative image features for predicting spatial transformations at higher spatial resolutions.

Similar to conventional multi-resolution image registration algorithms, the similarity of registered images at different resolutions is maximized in our network to serve as deep supervision [15], but without the need of supervised deformation field information. Such a supervised learning with surrogate supervised information is referred to as self-supervision in our study.

Different from conventional multi-resolution image registration algorithms in which deformation fields at lower-resolutions are typically used as initialization input to image registration at a higher spatial resolution, our deep learning based method jointly optimizes deformation fields at all resolutions with a typical feedforward and backpropagation based deep learning setting. As the optimization of the loss function proceeds, the parameters within the network will be updated through the feedforward computation and backpropagation procedure, leading to improved prediction of deformation fields. It is worth noting that no training deformation field information is needed for the optimization, and self-supervision through maximizing image similarity with smoothness regularization of deformation fields is the only force to drive the optimization.

**Network training**

Given a set of $n$ images, we could obtain $n^2$ pairs of fixed and moving images so that every image can serve as a fixed image. Pairs of images are registered using following parameters. In the proposed network, as illustrated in Fig. 1b, 32, 64, 128, and 64 kernels are used for Conv layer 1, 2, 3, and 4 respectively, with kernel size 3 and stride 2. For pooling layers, kernel size is set to 3, and stride 2. 64 and 32 kernels are used for Deconv layer 1 and 2 respectively, with kernel size 3 and stride 2. 3 kernels are used for regression layers 1, 2 and 3 to obtain 3D deformation fields. The total loss of the network is calculated as a weighted sum of loss 1 to 3, with weights coefficients as 1, 0.6, and 0.3. Alternative network architecture without pooling layers is implemented for performance comparison. Particularly, Conv modules 1 to 3, one regression layer, and one loss layer are kept. The Conv layers has the same parameters as described above. Moreover, alternative network architecture with pooling layers and additional one interpolation layer is also implemented as the coarse-grained registration model, tri-linear interpolation is adopted to upsample the coarse-grained deformation fields to the original spatial resolution of the input images.

All the models are implemented using Tensorflow [16]. Adam optimization technique [17] was adopted to train the networks, with learning rate set to 0.001, and batch size 64 (32 for the network without pooling layer due to GPU memory limit). The networks were trained on one Nvidia Titan Xp GPU, and 10000 iterations were adopted for the training. Once the training procedure using 640000 pairs of images is finished, the trained FCNs could be directly used to register new images with feedforward computation.

## Results

### Image dataset

The data used in this study were obtained from the Alzheimer's Disease Neuroimaging Initiative (ADNI) database (http://adni.loni.usc.edu), consisting of baseline MRI scans of 1776 subjects from ADNI 1, Go and 2. The ADNI was launched in 2003 as a public-private partnership, led by Principal Investigator Michael W. Weiner, MD. The primary goal of ADNI has been to test whether serial magnetic resonance imaging (MRI), positron emission tomography (PET), other biological markers, and clinical and neuropsychological assessment can be combined to measure the progression of mild cognitive impairment (MCI) and early Alzheimer's disease (AD). For up-to-date information, see www.adni-info.org. In particular, baseline MRI data of 959 subjects were obtained from the ADNI Go and 2 as training data, and baseline MRI data of 817 subjects in shared collections of the ADNI 1 were used as testing data.

In this study, we evaluated our image registration method to register hippocampus images. In particular, T1 MRI scans of all the subjects were registered to the MNI space using affine registration, and then a 3D bounding box of size 32×48×48 was adopted to extract hippocampus regions from the T1 image for each subject, as similarly did in a hippocampus segmentation study [18]. In addition, 100 T1 images with hippocampus segmentation labels were obtained from a preliminary release of the EADC-ADNI harmonized segmentation protocol project (www.hippocampal-protocol.net) [19]. These images with hippocampus segmentation labels were used to evaluate image registration performance based on an overlapping metric between the hippocampus labels of registered images.

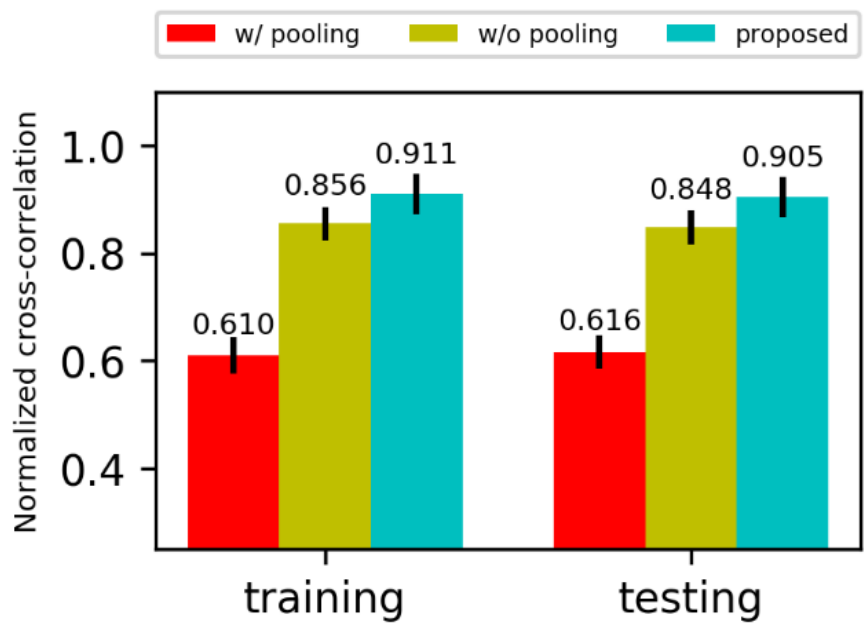

Fig. 2. Normalized cross-correlation of image pairs after registration from both training and testing datasets obtained by different deep learning models.

**Comparison with state-of-the-art image registration algorithms**

We compared our method with the deep learning methods that built upon CNNs in conjunction with STNs [7, 13] and ANTs [20] based on the same dataset. In particular, all the deep learning based image registration models were trained based the ADNI GO & 2 dataset, and evaluated based on the ADNI 1 dataset. 9600 pairs of images from ADNI 1 were randomly selected to test the performance of different deep learning models.

The image registration performance obtained by different deep learning models is illustrated in Fig. 2. The deep model without pooling layers, and deep model with multi-resolution strategy outperform the model with pooling layers significantly on both training and testing datasets, indicating that the coarse-grained deformation fields obtained could not capture fine-grained image correspondence. The proposed multi-resolution registration model obtained the best performance, demonstrating that the multi-resolution deeper model could extract more informative image feature representations and better learn spatial relationship between images. Interestingly, on both the training and testing datasets our deep learning models obtained similar image registration performance with respect to normalized cross-correlation between registered images.

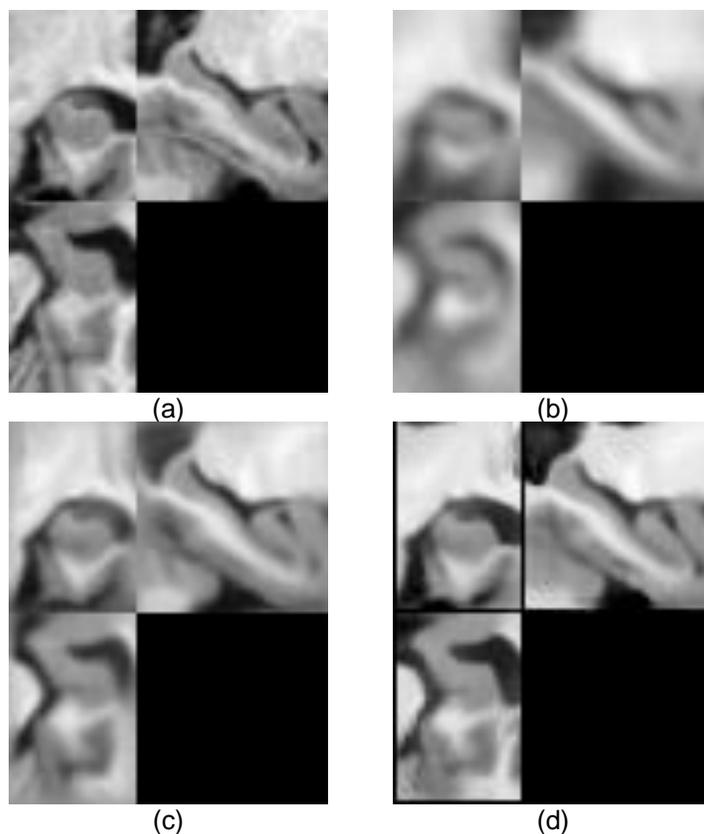

Fig. 3. Mean hippocampus image before and after registration by different methods. (a) Fixed image, (b) mean of images before registration, (c) mean of registered images by ANTs, (d) mean of registered image by the proposed method.

To compare our method with the ANTs algorithm, we randomly selected one image from the preliminary release of the EADC-ADNI harmonized segmentation protocol project as the fixed image, and registered all other images to the fixed image. For our method, we directly used the trained deep learning model to register these images, and ANTs registered the images with following command: ANTS 3 -m CC [fixed, moving, 1, 2] –o output -t SyN[0.25] –r Gauss[3,0] -i 100x100x10 --number-of-affine-iterations 100x100x100 --use-Histogram-Matching 1. The deformation fields obtained were applied to register their corresponding hippocampus label. The overlapping between the fixed and registered moving images was measured using $Dice = 2\frac{V(A \cap B)}{V(A)+V(B)}$, where $A$ is the hippocampus label of the fixed image, $B$ is the registered hippocampus label of the moving image, and $V(X)$ is volume of the hippocampus label X.

The mean image before and after registration by different methods are demonstrated in Fig. 3. As shown in Fig. 3b, the mean of images before registration is blurry. The means of registered images in Fig. 3c and 3d maintain detailed image textures, and the one obtained by the proposed method is sharper than that obtained by ANTs visually. Quantitatively, the overlap of hippocampus labels between images before registration was $0.654 \pm 0.062$, the overlap after registration by ANTs was $0.762 \pm 0.057$, and that by the proposed method was $0.798 \pm 0.033$. These results indicate that the proposed method could identify better spatial correspondence between images. Moreover, it took about 1 minute to register two images by ANTs on one CPU (AMD Opteron 4184 @ 2.80Ghz), while only about 50 ms by our trained deep model on one Titan Xp GPU.

## Discussions and Conclusion

We present a novel deep learning based non-rigid image registration algorithm to optimize and learn spatial transformations between pairs of images to be registered. Our deep learning model is built upon FCNs to facilitate voxel-to-voxel prediction of deformation fields at the same spatial resolution of images to be registered. The image registration is implemented in a multi-resolution image registration framework to jointly optimize and learn spatial transformations and FCNs at different resolutions with deep self-supervision through feedforward and backpropagation computation. The experimental results based on 3D structural MR images have demonstrated that our method could obtain promising image registration performance.

Different from most existing deep learning based image registration methods that learn spatial transformations for individual voxels or at a coarse-grained spatial resolution [7-10, 13], our method adopts FCNs with both convolutional and deconvolutional layers in addition to pooling layers to achieve voxel-to-voxel prediction of deformation fields at the same spatial resolution of images to be registered. The experimental results have demonstrated that our network architecture led to better image registration performance than alternatives, including CNNs for the prediction of coarse-grained deformation field and CNNs without pooling layers.

Different from those methods relying on training data with known deformation fields, our method directly estimates spatial transformations between pairs of images by maximizing an image-wise similarity metric between fixed and deformed moving images, similar to conventional image registration algorithms. Therefore, the registration of a set of images is also a training procedure for FCNs so that the trained FCNs can be directly adopted to register new images by feedforward computation of the learned FCNs. More importantly, the trained fully convolutional networks could be further improved for predicting spatial transformations between images if the optimization of spatial transformation is utilized to register new images with the backpropagation computation. The experimental results have demonstrated that the registration performance gauged with normalized correlation coefficients between registered images for both training and testing data was similar, indicating that the training procedure steered by image similarity based self-supervision is less prune to overfitting, unlike conventional supervised models.

Our deep learning based image registration model is implemented in a multi-resolution image registration framework, similar to conventional multi-resolution image registration algorithms [2]. However, the training for deep learning models with feedforward and backpropagation computation jointly optimizes deformation fields at different resolutions, fundamentally different from conventional multi-resolution image registration algorithms in which the computation of higher-resolution deformation fields relies on lower-resolution deformation fields, but could not provide any feedback to the computation of lower-resolution deformation fields. Such a joint optimization mechanism of the deep learning based image registration may facilitate better optimization of deformation fields at all different spatial resolutions. The experimental results have demonstrated that the deep leaning based image registration model could achieve better registration performance than ANTs with respect to overlapping of the registered hippocampus regions.

In conclusion, our study has demonstrated that deep learning models built upon fully convolutional networks with deep self-supervision could achieve promising performance for non-rigid image registration and would improve medical image registration performance with respect to both image registration accuracy and computational speed.